\documentclass[journal]{vgtc}                     

\vgtccategory{Research}

\vgtcpapertype{Algorithm}

\graphicspath{{figs/}{assets/}{figures/}{pictures/}{images/}{./}} 

\usepackage{tabu}                      
\usepackage{booktabs}                  
\usepackage{lipsum}                    
\usepackage{mwe}                       

\usepackage{mathptmx}                  
\usepackage{multirow}
\usepackage{enumitem}
\usepackage{comment}
\usepackage{cite}
\usepackage{lipsum} 
\usepackage{amsmath,amsfonts}
\usepackage{algorithmic}
\usepackage{algorithm}
\usepackage{array}
\usepackage[caption=false,font=normalsize,labelfont=sf,textfont=sf]{subfig}
\usepackage{textcomp}
\usepackage{stfloats}
\usepackage{url}
\usepackage{verbatim}
\usepackage{graphicx}
\usepackage{cite}
\usepackage{comment}
\usepackage{multirow}
\usepackage{amssymb}
\usepackage{booktabs}
\usepackage{cuted}
\usepackage{capt-of}
\usepackage{enumitem}

\def\shownotes{0}

\ifnum\shownotes=1
\newcommand\zhiming[1]{\textcolor{cyan}{Zhiming: #1}}
\newcommand\todo[1]{\textcolor{magenta}{#1}}
\newcommand\syn[1]{\textcolor{green}{Syn: #1}}
\newcommand\hl[1]{\textcolor{red}{#1}}
\newcommand\andreas[1]{\textcolor{orange}{Andreas: #1}}
\newcommand\daniel[1]{\textcolor{blue}{Daniel: #1}}
\else 
\newcommand\andreas[1]{}
\newcommand\syn[1]{}
\newcommand\zhiming[1]{}
\newcommand\daniel[1]{}
\newcommand\hl[1]{#1}
\newcommand\todo[1]{}
\fi

\newcommand{\methodName}{HOIMotion\xspace}

\title{\methodName: Forecasting Human Motion During Human-Object Interactions Using Egocentric 3D Object Bounding Boxes}

\author{%
  Zhiming Hu, Zheming Yin, Daniel Haeufle, Syn Schmitt, Andreas Bulling
}

\authorfooter{
  \item Zhiming Hu, Zheming Yin, Syn Schmitt, and Andreas Bulling are with the University of Stuttgart, Germany. E-mail: \{zhiming.hu@vis.uni-stuttgart.de, st178328@stud.uni-stuttgart.de, schmitt@simtech.uni-stuttgart.de, andreas.bulling@vis.uni-stuttgart.de\}.
   
   \item Daniel Haeufle is with Heidelberg University and University of Tuebingen, Germany. E-mail: daniel.haeufle@ziti.uni-heidelberg.de.

   \item Daniel Haeufle, Syn Schmitt, and Andreas Bulling are with the Center for Bionic Intelligence Tuebingen Stuttgart (BITS), Germany.
   
   \item Zhiming Hu is the corresponding author.
}

\abstract{%
  We present \textit{\methodName}~-- a novel
approach for human motion forecasting during human-object interactions that integrates information about past body poses and egocentric 3D object bounding boxes.
Human motion forecasting is important in many augmented reality applications
but most existing methods have only used past body poses to predict future motion.
\methodName first uses an encoder-residual graph convolutional network (GCN) and multi-layer perceptrons to extract features from body poses and egocentric 3D object bounding boxes, respectively.
Our method then fuses pose and object features into a novel pose-object graph and uses a residual-decoder GCN to forecast future body motion.
We extensively evaluate our method on the Aria digital twin (ADT) and MoGaze datasets and show that \methodName consistently outperforms state-of-the-art methods by a large margin of up to 8.7\% on ADT and 7.2\% on MoGaze in terms of mean per joint position error.
Complementing these evaluations, we report a human study (N=20) that shows that the improvements achieved by our method
result in forecasted poses being perceived as both more precise
and more realistic than those of existing methods.
Taken together, these results reveal the significant information content available in egocentric 3D object bounding boxes for human motion forecasting and the effectiveness of our method in exploiting this information.
}


\teaser{
  \centering
  \includegraphics[width=0.9\linewidth]{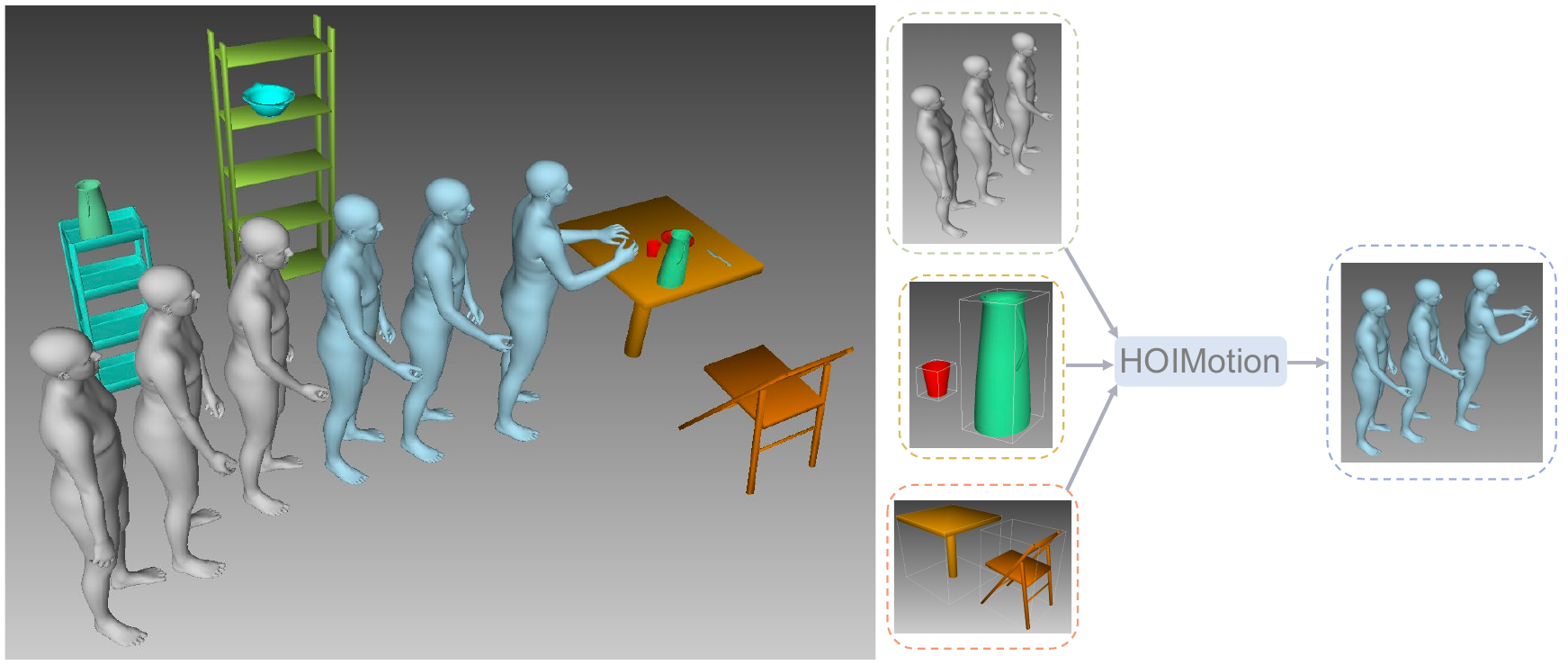}
  \caption{\methodName is a novel method for forecasting human motion during daily human-object interaction activities. The left figure shows an example of daily pick activity. \methodName uses body poses in the past and the 3D bounding boxes of scene objects in the egocentric view to forecast future human motion and can achieve superior performances over prior methods that only use historical body poses.
  }
  \label{fig:teaser}
}

\begin{document}


\maketitle

\section{Introduction}


Understanding and analysing human behaviour is a long-standing research challenge in virtual (VR) and augmented reality (AR) and is considered a crucial component for future human-aware intelligent VR/AR systems~\cite{hu2021fixationnet, hadnett2019effect}.
Human motion forecasting in particular
has significant relevance for a number of VR/AR applications including 1) redirected walking~\cite{cho2018path} that can redirect a user's walking path based on predicted future trajectories to create the illusion of an unlimited virtual interaction space; 2) collision avoidance~\cite{zheng2022gimo} that can avoid potential collision between two users or between a user and the physical world by predicting future human motion and sending out a warning beforehand if a collision is likely to happen; 3) low-latency interaction~\cite{david2021towards} that can prepare the virtual content in advance based on predicted future human motion to provide users with a low-latency experience; as well as  4) assistive devices~\cite{butaslac2020feasibility} that first predict users' desired future movements and then help them to accomplish them.

Human motion forecasting is typically formulated as a pose-focused sequence-to-sequence task, i.e. the task of using a sequence of past body poses to predict future poses.
This approach works reasonably well given that human behaviour, particularly during procedural or repetitive tasks, has rich internal structure and consists of characteristic, and thus predictable, sequences of actions.
Human motion, however, is also closely linked to the environment, especially during daily human-object interaction (HOI) activities.
For example, as illustrated in \autoref{fig:teaser}, if a user wants to pick up an object on the table, their body motion trajectories are strongly influenced by the location of the table while their arm movements are highly correlated with the specific position of the target object.
Inspired by the close link between human motion and scene objects, in this work we explore the potential of using information on scene objects to improve motion forecasting.
To ensure the generalisability in various VR/AR scenarios, we propose to only use egocentric 3D object bounding boxes since such information is readily available in VR/AR systems~\cite{park2008multiple, crivellaro2017robust, hu2021fixationnet}.


Information on human body pose and egocentric 3D object bounding box are so different that we cannot directly integrate them into existing methods, whose architectures are designed to only model human motion~\cite{ma2022progressively, guo2023back, mao2021multi, martinez2017human}.
To solve this problem, 
we present \textit{\methodName}~-- a novel graph convolutional network-based (GCN-based) encoder-residual-decoder architecture that can efficiently combine historical body poses and egocentric 3D object bounding boxes to forecast human motion in the future.
Our method first uses an encoder-residual GCN and multi-layer perceptrons (MLPs) to extract pose and object features respectively.
These features are fused into a novel pose-object graph and a residual-decoder GCN is applied to forecast future body motion from the pose-object graph.
We extensively evaluate our method at different future time horizons of up to $1$ second (future 30 frames) on the Aria digital twin (ADT) dataset~\cite{pan2023aria} for AR setting as well as on the MoGaze dataset~\cite{kratzer2020mogaze} for real-world setting.
Experimental results demonstrate that our method consistently outperforms several state-of-the-art methods that only use past body poses by a large margin, achieving an average improvement of 8.7\% on ADT and 7.2\% on MoGaze in terms of mean per joint position error (MPJPE).
Using only past body poses as input, our method can achieve an average improvement of 5.3\% on ADT and 3.7\% on MoGaze, validating the effectiveness of our architecture.
To qualitatively evaluate our method, we further conduct a user study and the responses from $20$ users validate that our predictions are perceived as both more precise and more realistic than predictions of prior methods.
The full source code and trained models are available at zhiminghu.net/hu24\_hoimotion.

\vspace{1em}
\noindent
The specific contributions of our work are three-fold:
\begin{itemize}
\item We demonstrate the effectiveness of egocentric 3D object bounding boxes for motion forecasting, providing a new perspective for this challenging task.

\item We propose a novel GCN-based encoder-residual-decoder architecture that uses an encoder-residual GCN and MLPs to extract pose and object features respectively, fuses these features into a pose-object graph, and applies a residual-decoder GCN to forecast future motion.

\item We report extensive experiments on two public datasets for forecasting human motion at different future time horizons and demonstrate significant performance improvements over state of the art and report a user study that shows our method achieves superior performances over prior methods in both precision and realism.
\end{itemize}
\section{Related Work}

\subsection{Human Behaviour Modelling}
Understanding and modelling human behaviours is a long-standing research challenge in the areas of virtual and augmented reality~\cite{hu24_pose2gaze, valkov2023reach, bektacs2023gear, razali2022using, adebayo2022hand, jiao23_supreyes} and is considered a significant component for future human-aware intelligent VR/AR systems~\cite{hu2021fixationnet, hadnett2019effect}.
Many researchers focused on visual attention modelling in virtual reality.
Specifically, Sitzmann et al. modelled human visual saliency on 360-degree VR images and proposed to combine human head orientation to predict saliency map in VR~\cite{sitzmann2018saliency}.
Hu et al. focused on human gaze behaviours in VR and proposed several eye-head coordination models to predict eye gaze positions in free-viewing~\cite{hu2019sgaze,hu2020dgaze} and task-oriented virtual environments~\cite{hu2021fixationnet}.
Some researchers concentrated on modelling human cognitive load in virtual environments~\cite{tremmel2019estimating, dell2020cognitive}.
For example, Tremmel et al. proposed to use electroencephalogram features to estimate cognitive load in an interactive virtual environment~\cite{tremmel2019estimating} while Dell'Agnola et al. extracted features from different physiological signals to detect the levels of cognitive load~\cite{dell2020cognitive}.
Hu et al. analysed different human activities in VR and proposed to use human eye and head movements to recognise user activities~\cite{hu2022ehtask}.
Kim et al. developed an electroencephalography-driven model to predict the degree of cybersickness in virtual environments~\cite{kim2019deep}.
In this work, we focused on human motion modelling, specifically predicting human full-body motion in the future.

\subsection{Human Motion Forecasting}
Human motion forecasting is a significant research topic in the area of human-centred computing and has great relevance for many VR/AR applications.
Early works usually employed traditional machine learning methods to model human motion.
Specifically, Wang et al. employed Gaussian process models to learn an effective representation of human motion data~\cite{wang2005gaussian}, Taylor et al. used a restricted Boltzmann machine to model the probability distribution of human body poses~\cite{taylor2006modeling}, while Lehrmann et al. proposed to model human motion through hidden Markov models~\cite{lehrmann2014efficient}.
While these early methods are useful for simple motions, they are less effective for forecasting more complex and long-term motion sequences~\cite{fragkiadaki2015recurrent}.
Recently, with the rapid development of deep learning technology, many deep learning-based methods have been proposed to model human body motion.
Considering the sequential structure of human motion, many researchers have explored to forecast human future motion using recurrent neural networks (RNNs) and have achieved superior performances over traditional methods~\cite{le2021hierarchical, fragkiadaki2015recurrent, martinez2017human}.
In addition to RNNs, Transformers have also been applied to this task and have achieved good results~\cite{mao2021multi, aksan2021spatio}.
More recently, some researchers explored to forecast human motion using graph convolutional networks in light of the fact that human body pose can be viewed as a graph by treating each body joint as a graph node~\cite{ma2022progressively, dang2021msr}.
To reduce the network complexity, multi-layer perceptrons have been proposed as a light-weight motion forecasting solution~\cite{guo2023back}.
Existing motion forecasting methods typically only focused on human motion itself, forecasting future body poses using only historical poses.
Recent work on offline human motion synthesis used the features from a global 3D scene point cloud to synthesise human motion~\cite{zheng2022gimo, mao2022contact}.
However, such features are difficult to acquire in many real application scenarios, especially in augmented or mixed reality settings, thus limiting the usefulness of these methods in real-day life outside clearly defined environments. 
In contrast with previous work, in this work we focus on real-time motion forecasting and combine body poses in the past and information on egocentric 3D object bounding boxes, which is readily available in VR/AR systems, to predict human motion in the future.



\subsection{Human-object Interaction}
Human-object interaction is a crucial interaction paradigm in virtual and augmented reality~\cite{hasson2019learning, ohkawa2023assemblyhands, hu2020gaze, hu2021eye}.
Recent research has revealed the strong correlation between human behaviours and the scene objects during daily human-object interaction activities.
Specifically, Hu et al. studied the visual search setting where users were required to search for a specific target object among many distractors and found that both the target and distractors have a strong influence on human gaze behaviours~\cite{hu2021fixationnet}.
Li et al. revealed that users' spatial memory of the scene content influences their visual search strategies in large-scale immersive virtual environments~\cite{li2018memory}.
David-John et al. found that during an item-selection activity in virtual reality human eye gaze is closely linked with the items that users intend to select~\cite{david2021towards}.
Koulieris et al. revealed that during the process of game play, player actions are highly correlated with the present states of the game-related objects~\cite{koulieris2016gaze}.
Emery et al. investigated open-ended VR games that covered various HOI activities such as shooting and object manipulation and revealed that human eye, head, and hand movements are strongly linked with the scene objects~\cite{emery2021openneeds}.
Inspired by the close link between human behaviours and scene objects, in this work we introduce to use information on scene objects to forecast human motion during human-object interactions.

\section{Method}

\begin{figure*}
\centering
    \includegraphics[width=1.0\textwidth]{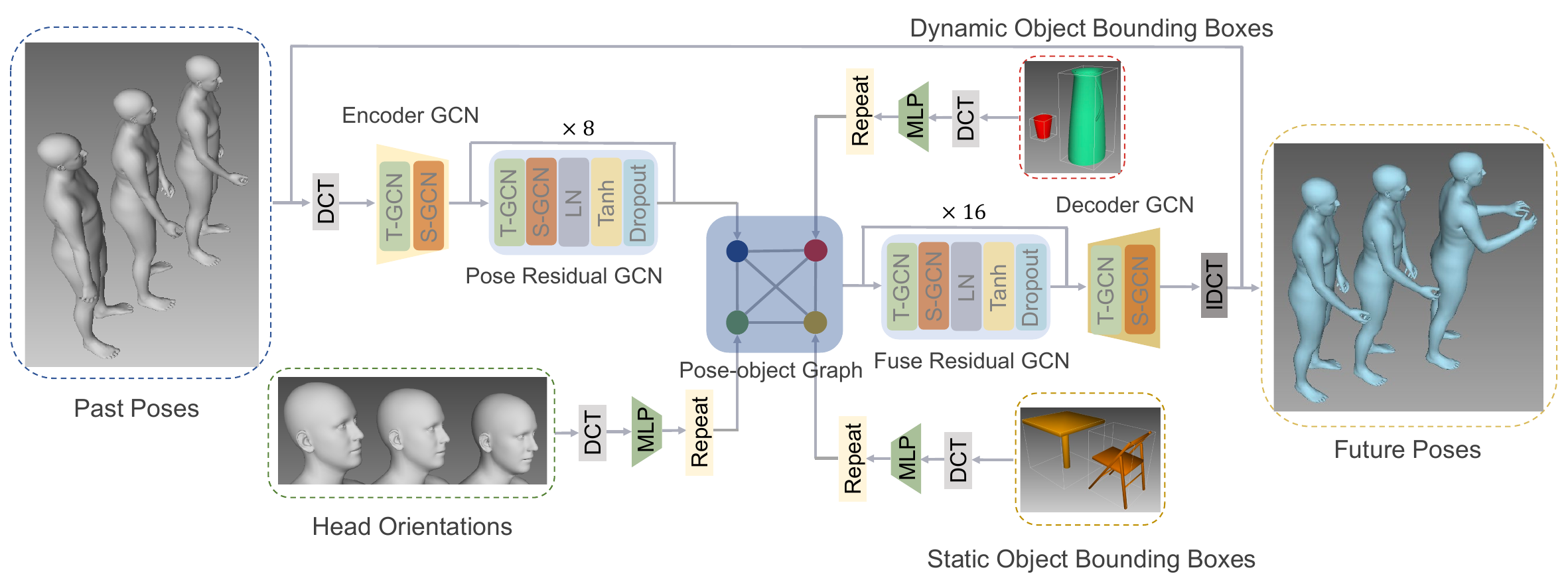}
    \caption{Our method for egocentric scene object-aware human motion forecasting uses an encoder GCN and a pose residual GCN to extract features from historical body poses and employs three MLPs to respectively extract features from head orientations, and the bounding boxes of static and dynamic scene objects in the egocentric view.
    The pose, head, and object features are fused into a novel pose-object graph and a fuse residual GCN and a decoder GCN are applied to forecast future body motion from the pose-object graph.}
    \label{fig:method}
\end{figure*}

\subsection{Problem Definition}

We define egocentric scene object-aware human motion forecasting as the task of predicting a sequence of future body poses jointly from historical body poses and information on scene objects in the egocentric view.
We use the 3D positions of all human body joints to represent body pose $p\in R^{3\times n}$, where $n$ is the number of joints.
We represent scene objects using their bounding boxes given that the bounding box information can be easily and efficiently accessed in VR/AR systems and is highly relevant to human motion.
Specifically, we use the 3D positions of the bounding box's eight vertices to represent scene objects $o\in R^{3\times 8\times m}$, where $m$ is the number of objects.
Considering that a user's egocentric viewport is determined by their head orientation in VR/AR systems, we also introduce to use human head orientation $h\in R^3$ in this task, where $h$ is a unit vector indicating head forward direction.
Given a sequence of historical body poses $P_{1:t} = \{p_1, p_2, ..., p_t\}$, head orientations $H_{1:t} = \{h_1, h_2, ..., h_t\}$, and scene objects $O_{1:t} = \{o_1, o_2, ..., o_t\}$, the task is to forecast body poses in the future $P_{t+1:T} = \{p_{t+1}, p_{t+2}, ..., p_{T}\}$.
The core of our method is a GCN-based encoder-residual-decoder architecture. 
An encoder GCN and a pose residual GCN are used to extract features from historical body poses while MLPs are employed to extract features from head orientations and scene objects.
The pose, head, and object features are fused into a novel pose-object graph, and a fuse residual GCN and a decoder GCN are applied to forecast future body motion from the pose-object graph (see \autoref{fig:method} for an overview of our method).


\subsection{Pose Feature Extraction} \label{sec:pose_feature}

To cover the future time horizon that we want to predict, we first padded the historical body poses $P_{1:t}\in R^{3\times n \times t}$ from a temporal length of $t$ to $T$ by repeating the last observed pose $p_t$ for $T-t$ times following prior works~\cite{ma2022progressively, mao2019learning}.
We applied discrete cosine transform (DCT) to encode the padded body poses $P\in R^{3\times n \times T}$ in the temporal domain in light of the good performance of DCT for encoding time series data~\cite{ma2022progressively,mao2020history}:
\begin{equation} 
\begin{aligned}
    P_{dct} = P M_{dct}, 
\end{aligned}
\end{equation}
where $M_{dct} \in R^{T\times T}$ is the DCT matrix and $P_{dct}\in R^{3\times n \times T}$ is the transformed body poses.
We further proposed two GCN blocks, i.e. an encoder GCN block and a pose residual GCN block, to extract features from the transformed body pose data.

\paragraph{Encoder GCN}
The encoder GCN block aims at mapping the body pose data from its original space into a latent feature space.
This block first employs a temporal GCN (T-GCN) to extract temporal features from the transformed pose data $P_{dct}$.
The T-GCN views the body pose data as a fully-connected temporal graph that contains $T$ nodes corresponding to $T$ time steps of the pose data.
The core of the T-GCN is a weighted adjacency matrix $A^T \in R^{T\times T}$ that learns the correlations between different temporal nodes and performs temporal convolution on the pose data:
\begin{equation} 
\begin{aligned}
    f_{temp} = P_{dct} A^T.
\end{aligned}
\end{equation}
$f_{temp}\in R^{3\times n\times T}$ was then permuted to $f_{temp}\in R^{T\times n\times 3}$ and a weight matrix $W^{start}\in R^{3\times16}$ was used to map the original node features ($3$ dimensions) into latent space ($16$ dimensions):
\begin{equation} 
\begin{aligned}
    f_{lat} = f_{temp} W^{start},
\end{aligned}
\end{equation}
where $f_{lat}\in R^{T\times n\times 16}$ represents the latent features.
After the weight matrix, a spatial GCN (S-GCN) was employed to extract spatial features from the pose data.
The S-GCN views the body pose data as a full-connected spatial graph that contains $n$ nodes corresponding to $n$ human body joints.
The core of the S-GCN is a weighted adjacency matrix $A^S \in R^{n\times n}$ that learns the link between different spatial nodes and performs spatial convolution on the pose data:
\begin{equation} 
\begin{aligned}
    f_{spa} = A^S f_{lat}.
\end{aligned}
\end{equation}
$f_{spa}\in R^{T\times n\times 16}$ was further permuted to $f_{spa}\in R^{16\times n\times T}$ for further processing.

\paragraph{Pose Residual GCN}
The pose residual GCN block is designed to further enhance the body pose features.
It first copies the body pose features along the temporal dimension ($R^{16\times n\times T}\rightarrow R^{16\times n\times 2T}$) to enhance the temporal features~\cite{ma2022progressively}.
It then applies eight GCN components to further process the body pose data.
Each GCN component contains a temporal GCN that learns the temporal features using the temporal adjacency matrix $A^T \in R^{2T\times 2T}$, a weight matrix $W^{res}\in R^{16\times16}$ that learns latent features, a spatial GCN that extracts spatial features through the spatial adjacency matrix $A^S \in R^{n\times n}$, a layer normalisation (LN), a Tanh activation function, as well as a dropout layer with a dropout rate of $0.3$ to prevent the GCN from overfitting.
A residual connection was added for each GCN component to improve the network flow.
The output of the pose residual GCN block was cut in half in the temporal dimension to obtain the body pose features $f_{pose}\in R^{16\times n\times T}$ that maintain the same temporal length as the original input to this block.

\subsection{Head Feature Extraction}
We first padded the historical head orientations $H_{1:t}\in R^{3\times t}$ to a temporal length of $T$ by repeating the last observed head orientation $h_t$ for $T-t$ times.
A DCT is then applied to encode the head orientations in the temporal domain.
We further used a multi-layer perceptron to process the head orientations given the effectiveness of MLP for encoding the sequences of human behaviour data~\cite{hu2022ehtask, hu2020dgaze}.
The MLP aims at mapping the head features ($3$ dimensions) at each time step into latent space ($16$ dimensions).
Specifically, we used three linear layers with $128$, $128$, $16$ neurons respectively to extract features from head orientations.
The first two linear layers were followed by a layer normalisation, a Tanh activation function, as well as a dropout layer with a dropout rate of $0.5$ to avoid overfitting while a layer normalisation and a Tanh activation function were employed after the third linear layer.
Through the multi-layer perceptron, we obtained the processed head orientation features $f_{head}\in R^{16\times T}$.

\subsection{Object Feature Extraction}\label{sec:object_feature}

Considering that different type of objects may have different influences on human motion during human-object interaction activities, we grouped the scene objects into two categories, i.e. \textit{dynamic} and \textit{static} objects, following prior work on object segmentation~\cite{pan2023aria}.
Dynamic objects refer to the objects that users can manipulate to change their positions during an HOI activity, e.g. the cup and jug in \autoref{fig:teaser}, while static objects denote the objects that are stationary throughout the HOI activity, e.g. the table and chair in \autoref{fig:teaser}.
An environment may contain plenty of scene objects, many of which may have little influence on human body motion.
To improve our method's efficiency, we only used the scene objects that were located in the central region of users' viewport since these central objects are more likely to influence human behaviours~\cite{sitzmann2018saliency,hu2019sgaze,hu2020dgaze}.
Specifically, at each time step we calculated the angular distance between the centre of each scene object and the centre of the viewport and ranked the dynamic and static objects respectively based on their angular distances from the viewport centre.
We then selected the two dynamic objects $D_{1:t} = \{d_1, d_2, ..., d_t\}\in R^{3\times 8\times 2 \times t}$ and two static objects $S_{1:t} = \{s_1, s_2, ..., s_t\}\in R^{3\times 8\times 2 \times t}$ that are closest to the viewport centre and used their bounding box information for motion forecasting.

To cover the predicted future time horizon, we padded the object information to a temporal length of $T$ by respectively repeating the last observed objects $d_t$ and $s_t$ for $T-t$ times.
We then applied a DCT to encode the scene objects in the temporal domain and further used two MLPs to extract features from dynamic and static objects respectively.
The two MLPs have the same structure and are designed to map the object features ($3\times8\times2$ dimensions) at each time step into latent space ($16$ dimensions).
Each MLP contains three linear layers with $128$, $128$, $16$ neurons respectively.
A layer normalisation, a Tanh activation function, and a dropout layer with a dropout rate of $0.5$ were applied after the first two linear layers while the third linear layer was followed by a layer normalisation and a Tanh activation function.
Through the two MLPs, we obtained the processed features of both dynamic objects $f_{dynamic}\in R^{16\times T}$ and static objects $f_{static}\in R^{16\times T}$.

\subsection{Pose-object Fusion} \label{sec:fusion}
After the feature extraction process, we obtained the body pose features $f_{pose}\in R^{16\times n\times T}$, head orientation features $f_{head}\in R^{16\times T}$, dynamic object features $f_{dynamic}\in R^{16\times T}$, as well as static object features $f_{static}\in R^{16\times T}$.
To enhance the head and object features,
we respectively repeated the head and object features for four times along the spatial domain and obtained $f_{head}\in R^{16\times 5\times T}$, $f_{dynamic}\in R^{16\times 5\times T}$, and $f_{static}\in R^{16\times 5\times T}$.
We then concatenated the pose, head, and object features along the spatial domain and obtained $f\in R^{16\times (n+15)\times T}$.
To fuse different features for motion forecasting, we proposed a novel spatio-temporal pose-object graph: the temporal graph covers $T$ nodes that correspond to the features at $T$ time steps while the spatial graph contains $n+15$ nodes corresponding to the features of the body joints ($n$ nodes), head orientations ($5$ nodes), dynamic objects ($5$ nodes), and static objects ($5$ nodes).
Both the temporal and spatial graphs are fully-connected with their adjacency matrices measuring the weights between each pair of nodes.

\subsection{Motion Forecasting}
We further employed a fuse residual GCN block and an end GCN block to forecast future body poses from the pose-object graph.

\paragraph{Fuse Residual GCN}
The fuse residual GCN block aims at enhancing the fused pose-object features.
It first copies the pose-object features along the temporal dimension ($R^{16\times(n+15)\times T}\rightarrow R^{16\times (n+15)\times 2T}$) to enhance the temporal features and then applies $16$ GCN components to further process the pose-object data.
Each GCN component consists of a temporal GCN that learns the temporal features using the temporal adjacency matrix $A^T \in R^{2T\times 2T}$, a weight matrix $W^{res}\in R^{16\times16}$ that learns latent features, a spatial GCN that extracts spatial features through the spatial adjacency matrix $A^S \in R^{(n+15)\times (n+15)}$, a layer normalisation, a Tanh activation function, and a dropout layer with a dropout rate of $0.3$ to avoid overfitting.
A residual connection was added for each GCN component to improve the network flow.
We cut the output of the fuse residual GCN block in half in the temporal dimension to maintain the same temporal length as the original input and obtained the spatio-temporal pose-object features $f\in R^{16\times (n+15)\times T}$.

\paragraph{Decoder GCN} 
The decoder GCN block is employed to map the processed pose-object features from latent feature space to the original space.
The decoder GCN consists of a temporal GCN that learns the temporal adjacency matrix, a weight matrix $W^{end}\in R^{16\times3}$ that maps the latent features to three dimensions, and a spatial GCN that learns the spatial adjacency matrix.
The output of the decoder GCN $Y_d \in R^{3\times (n+15)\times T}$ was converted back to the original representation space using an inverse discrete cosine transform (IDCT) matrix $M_{idct} \in R^{T\times T}$:
\begin{equation} 
\begin{aligned}
    Y = Y_dM_{idct}.
\end{aligned}
\end{equation}
We finally added a global residual connection between the pose input and the output of IDCT to obtain the predicted future poses $\hat{P}_{t+1:T} \in R^{3\times n\times {T-t}}$.



\subsection{Loss Function}

To ensure the precision and smoothness of our predictions, we employed a combination of motion loss $\ell_m$ and velocity loss $\ell_v$ as our loss function $\ell$:
\begin{equation} 
\begin{aligned}
    \ell = \ell_m + \ell_v.
\end{aligned}
\end{equation}
$\ell_m$ is designed to measure our method's precision by calculating the mean per joint position error between the ground truth and the predicted future poses~\cite{ma2022progressively, mao2020history}:
\begin{equation} 
\begin{aligned}
\ell_m =\frac{1}{n(T-t)}\sum_{i=t+1}^T\sum_{j=1}^n\lVert p_{i,j} - \hat{p}_{i,j} \rVert^2,
\end{aligned}\label{eq:mpjpe}
\end{equation}
where $p_{i,j} \in R^3$ represents the ground truth 3D coordinates of the $j^{th}$ joint at the future time of $i$ while $\hat{p}_{i,j} \in R^3$ is the prediction of our method.
$\ell_v$ aims at measuring the smoothness of our predictions by computing the mean per joint velocity error between the ground truth and the predicted future poses~\cite{guo2023back}:
\begin{equation} 
\begin{aligned}
\ell_v =\frac{1}{n(T-t-1)}\sum_{i=t+1}^{T-1}\sum_{j=1}^n\lVert v_{i,j} - \hat{v}_{i,j}\rVert^2,
\end{aligned}\label{eq:vel_loss}
\end{equation}
where $v_{i,j} \in R^3$ is the ground truth pose velocity and $\hat{v}_{i,j} \in R^3$ represents the predicted pose velocity.
The velocity is computed using the difference between two adjacent poses: $v_{i,j} = p_{i+1,j} - p_{i,j}$ and $\hat{v}_{i,j} = \hat{p}_{i+1,j} - \hat{p}_{i,j}$.
\section{Experiments and Results}
In this section, we conducted extensive experiments to evaluate our method's motion forecasting performance.
Specifically, we first compared our method with the state-of-the-art methods that only use historical body poses on an AR dataset as well as on a real-world dataset.
We further performed extensive ablation studies to validate the effectiveness of each component used in our method.
We finally conducted a user study to qualitatively evaluate our method.

\begin{table*}[h]
        \caption{Mean per joint position errors (unit: millimeters) of different methods for motion forecasting on the ADT and MoGaze datasets. Results are shown for different future time horizons of up to 1 second. Best results are in bold. Our method consistently outperforms prior methods in terms of average performance as well as performances at different time intervals. Even using only historical body poses as input, our method still achieves significantly better performances over prior methods.}\label{tab:global}
	\centering
	\resizebox{0.9\textwidth}{!}{
	\begin{tabular}{ccccccccccccccccc}
		\toprule
		Action& Method & 100 ms & 200 ms & 300 ms & 400 ms & 500 ms & 600 ms & 700 ms & 800 ms & 900 ms & 1000 ms &Average\\ \hline
	\multirow{6}{*}{\textit{ADT-work}}
        &\textit{Res-RNN}~\cite{martinez2017human} &28.4 &38.7 &47.8 &58.1 &67.3 &76.7 &85.9 &95.1 &104.2 &113.1 &69.1\\
        &\textit{siMLPe}~\cite{guo2023back} &26.0 &28.4 &33.2 &39.8 &47.9 &55.1 &62.5 &71.4	&79.4 &89.4 &51.2\\
        &\textit{HisRep}~\cite{mao2021multi} &6.9 &12.8 &18.8 &24.8 &31.0 &37.4 &44.1 &50.9	&57.8 &65.3 &32.8\\
        &\textit{PGBIG}~\cite{ma2022progressively} &8.2 &13.5 &19.2 &24.9 &30.9 &37.2 &43.6	&50.1 &57.0 &64.5 &32.9\\
        &Ours \textit{pose only} &5.1 &10.4 &16.1 &22.2 &28.6 &35.3 &42.3	&49.5 &56.9 &64.6 &30.9\\
        &Ours &\textbf{4.9} &\textbf{10.0} &\textbf{15.6} &\textbf{21.6} &\textbf{27.8} &\textbf{34.3} &\textbf{41.1} &\textbf{48.0} &\textbf{55.2} &\textbf{62.5} &\textbf{30.0}\\
        \midrule
	\multirow{6}{*}{\textit{ADT-decoration}}
        &\textit{Res-RNN}~\cite{martinez2017human} &22.5 &33.4 &46.2 &59.7 &73.5 &87.4 &101.5	&115.6 &129.8 &144.2 &77.4\\
        &\textit{siMLPe}~\cite{guo2023back} &27.2 &33.6 &44.6 &56.7 &70.6 &85.4 &101.2 &118.5	&136.9 &156.6 &78.7\\
        &\textit{HisRep}~\cite{mao2021multi} &10.5 &19.2 &27.9 &37.2 &47.5 &58.8 &71.3 &84.9	&99.0 &114.4 &53.2\\
        &\textit{PGBIG}~\cite{ma2022progressively} &10.2 &18.7 &27.0 &35.7 &45.5 &56.5 &68.6 &81.9 &96.2 &111.7 &51.5\\
        &Ours \textit{pose only} &6.9 &14.2 &22.5 &32.0 &42.6 &54.4 &67.2 &81.1 &95.6 &110.9 &49.0\\
        &Ours &\textbf{6.6} &\textbf{13.5} &\textbf{21.5} &\textbf{30.6} &\textbf{40.8} &\textbf{52.2} &\textbf{64.7} &\textbf{77.8} &\textbf{91.6} &\textbf{105.7} &\textbf{46.9}\\
        \midrule
	\multirow{6}{*}{\textit{ADT-meal}}
        &\textit{Res-RNN}~\cite{martinez2017human} &18.7 &27.8 &39.3 &52.0 &65.4 &79.6 &94.5	&110.0 &125.8 &141.9 &71.5\\
        &\textit{siMLPe}~\cite{guo2023back} &26.7 &29.9 &37.0 &46.2 &57.1 &68.4 &81.1 &94.9	&108.2 &122.7 &63.9\\
        &\textit{HisRep}~\cite{mao2021multi} &8.0 &15.0 &22.4 &30.4 &39.1 &48.6 &58.6 &69.2 &80.0 &91.3 &43.2\\
        &\textit{PGBIG}~\cite{ma2022progressively} &8.5 &15.2 &22.2 &29.7 &38.0 &47.1 &56.8 &67.0 &77.9 &89.7 &42.3\\
        &Ours \textit{pose only} &5.6 &11.6 &18.6 &26.6 &35.4 &44.9 &55.2	&66.1 &77.4 &89.0 &40.0\\
        &Ours &\textbf{5.3} &\textbf{11.2} &\textbf{18.0} &\textbf{25.8} &\textbf{34.4} &\textbf{43.8} &\textbf{53.8} &\textbf{64.2} &\textbf{75.1} &\textbf{86.2} &\textbf{38.9}\\
        \midrule
	\multirow{6}{*}{\textit{ADT-all}}
        &\textit{Res-RNN}~\cite{martinez2017human} &23.7 &33.9 &44.8 &56.8 &68.6 &80.8 &93.1	&105.7 &118.3 &131.1 &72.3\\
        &\textit{siMLPe}~\cite{guo2023back} &26.6 &30.4 &37.8 &46.8 &57.5 &68.2 &79.7 &92.5	&105.3 &119.5 &63.2\\        
        &\textit{HisRep}~\cite{mao2021multi} &8.3 &15.4 &22.6 &30.2 &38.4 &47.2 &56.6 &66.6	&76.8 &87.8 &42.0\\
        &\textit{PGBIG}~\cite{ma2022progressively} &8.9 &15.5 &22.4 &29.6 &37.4 &46.0 &55.0 &64.7 &75.0 &86.2 &41.3\\
        &Ours \textit{pose only} &5.8 &11.9 &18.8 &26.4 &34.8 &43.9 &53.6	&63.9 &74.7 &85.8 &39.1\\
        &Ours &\textbf{5.5} &\textbf{11.4} &\textbf{18.1} &\textbf{25.6} &\textbf{33.7} &\textbf{42.5} &\textbf{52.0} &\textbf{61.8} &\textbf{72.0} &\textbf{82.5} &\textbf{37.7}\\
        \midrule   
        \midrule        
        \multirow{6}{*}{\textit{MoGaze-pick}}
        &\textit{Res-RNN}~\cite{martinez2017human} &36.1 &50.1 &67.9 &88.1 &110.6 &135.1	&161.3 &189.1 &218.1 &248.0 &123.6\\
        &\textit{siMLPe}~\cite{guo2023back} &27.4 &37.7 &51.5 &67.4 &84.7 &103.8 &125.1 &147.4 &171.2	&196.1 &95.4\\
        &\textit{HisRep}~\cite{mao2021multi} &14.7 &27.7 &41.1 &55.8	&72.1 &90.1	&109.6 &130.3 &151.9 &174.1 &80.9\\
        &\textit{PGBIG}~\cite{ma2022progressively} &13.9 &26.3 &39.2 &53.3 &69.4 &87.0 &105.8 &125.8 &146.8	&168.0 &78.0 \\
        &Ours \textit{pose only} &12.2 &23.7 &36.4 &50.4 &66.2 &83.6 &102.4	&122.0 &142.5 &163.4 &74.8\\
        &Ours &\textbf{11.3} &\textbf{22.6} &\textbf{34.9} &\textbf{48.8} &\textbf{64.3} &\textbf{81.3} &\textbf{99.8} &\textbf{119.1} &\textbf{139.1} &\textbf{159.3} &\textbf{72.7}\\
        \midrule
	\multirow{6}{*}{\textit{MoGaze-place}}
        &\textit{Res-RNN}~\cite{martinez2017human} &42.9 &58.6 &77.1 &97.0 &118.1 &140.1 &162.4 &184.5 &206.4	&227.6 &125.6\\
        &\textit{siMLPe}~\cite{guo2023back} &31.3 &45.9 &62.8 &80.3 &98.3 &118.0 &139.6 &162.1 &185.5 &210.0 &107.1\\
        &\textit{HisRep}~\cite{mao2021multi} &21.6 &38.3 &53.4 &69.3 &86.5 &105.2 &124.8 &144.6 &164.5 &184.9 &93.2\\
        &\textit{PGBIG}~\cite{ma2022progressively} &19.9 &35.2	&50.1 &65.8	&82.8 &101.1 &120.1 &139.6 &159.0 &177.9 &89.3\\
        &Ours \textit{pose only} &18.0 &32.8 &47.7 &63.4 &80.4 &98.4 &117.0 &135.9 &155.2 &174.4 &86.6\\
        &Ours &\textbf{16.7} &\textbf{31.1} &\textbf{45.6} &\textbf{60.9} &\textbf{77.1} &\textbf{94.2} &\textbf{111.9} &\textbf{129.9}	&\textbf{148.0} &\textbf{165.5} &\textbf{82.6}\\
        \midrule 
        \multirow{6}{*}{\textit{MoGaze-all}}
        &\textit{Res-RNN}~\cite{martinez2017human} &38.5 &53.1 &71.1 &91.3	&113.2 &136.8 &161.7 &187.5 &214.0 &240.8 &124.3\\
        &\textit{siMLPe}~\cite{guo2023back} &28.8 &40.6 &55.5 &72.0 &89.4 &108.8 &130.2	&152.6 &176.3 &201.0 &99.5\\
        &\textit{HisRep}~\cite{mao2021multi} &17.1 &31.4 &45.4 &60.5	&77.1 &95.4	&115.0 &135.3 &156.4 &177.9 &85.3\\
        &\textit{PGBIG}~\cite{ma2022progressively} &16.0 &29.4	&43.0 &57.7 &74.1 &92.0 &110.8 &130.7 &151.1 &171.5 &82.0\\ 
        &Ours \textit{pose only} &14.3 &26.9 &40.4 &55.0 &71.2 &88.8 &107.5	&126.9 &147.0 &167.3 &79.0\\
        &Ours &\textbf{13.2} &\textbf{25.6} &\textbf{38.6} &\textbf{52.9} &\textbf{68.7} &\textbf{85.7} &\textbf{103.9} &\textbf{122.7}	&\textbf{142.0} &\textbf{161.3} &\textbf{76.1}\\
        \bottomrule        
	\end{tabular}} 
\end{table*}

\subsection{Datasets}
To test our method's generalisation capability for different settings, we evaluated our method on two public datasets including an AR dataset (Aria digital twin~\cite{pan2023aria}) and a real-world dataset (MoGaze~\cite{kratzer2020mogaze}).

\paragraph{ADT dataset~\cite{pan2023aria}}
The Aria digital twin dataset is collected in two indoor environments (an apartment and an office environment) with simulated scene objects and contains human pose data performing various human-object interaction activities including \textit{room decoration}, \textit{meal preparation}, and \textit{work}.
Each human pose consists of the 3D coordinates of $21$ human joints recorded at $30$ Hz.
The bounding box information and motion type (\textit{dynamic} or \textit{static}) of the scene objects are also provided.
For experiments on ADT, we randomly selected $24$ sequences for training and $10$ sequences for testing.

\paragraph{MoGaze dataset~\cite{kratzer2020mogaze}}
The MoGaze dataset is collected in a real-world indoor environment and contains human motion data recorded at $120$ Hz from six people performing daily \textit{pick} and \textit{place} activities.
The bounding box information and motion type (\textit{dynamic} or \textit{static}) of the scene objects are also recorded at $120$ Hz.
We down-sampled the human pose and object data to $30$ Hz for simplicity~\cite{ma2022progressively, guo2023back} and represented human poses using the 3D coordinates of $21$ human joints.
To evaluate motion forecasting on MoGaze, we used a leave-one-person-out cross-validation: \hl{We trained on five participants from scratch and tested on the remaining one}, repeated the experiment six times with a different participant for testing, and calculated the average performance across all six iterations.

\subsection{Evaluation Settings}
\paragraph{Evaluation Metric} 
As is common in human motion forecasting~\cite{zheng2022gimo,guo2023back,ma2022progressively}, we used the mean per joint position error (see \autoref{eq:mpjpe}) in millimeters as our metric to evaluate different motion forecasting methods.

\paragraph{Baselines} 
\hl{We compared our method with the following methods because they are not only prior state-of-the-art motion forecasting methods but also representatives of different network architectures, i.e. RNN, MLP, Transformer, and GCN:} 
\begin{itemize}[noitemsep,leftmargin=*]
    
    \item \textit{Res-RNN}~\cite{martinez2017human}: \textit{Res-RNN} is a RNN-based method that applies a residual connection between the input pose and output pose to improve performance.

    \item \textit{siMLPe}~\cite{guo2023back}: \textit{siMLPe} is a light-weight MLP-based method that applies discrete cosine transform and residual connections to improve performance.
    
    \item \textit{HisRep}~\cite{mao2021multi}: \textit{HisRep} is a Transformer-based method that extracts motion attention to capture the similarity between the current motion context and the historical motion sub-sequences.
    
    \item \textit{PGBIG}~\cite{ma2022progressively}: \textit{PGBIG} is a GCN-based method that employs a multi-stage framework to forecast human motions where each stage predicts an initial guess for the next stage.
\end{itemize}

\paragraph{Time Horizons of Input and Output Sequences}
For experiments on the ADT and MoGaze datasets (30 Hz), we used $10$ frames of data as input to forecast human poses in the future $30$ frames (i.e. up to one second into the future), following the common evaluation settings for motion forecasting~\cite{ma2022progressively, mao2021multi}.

\paragraph{Implementation Details}
We trained the baseline methods from scratch using their default parameters.
For our motion forecasting network, we used the Adam optimiser with an initial learning rate of $0.01$ and decayed the learning rate by $0.95$ every epoch.
A batch size of $32$ was employed to train the motion forecasting network for $80$ epochs.
Our method was implemented using the PyTorch framework.

\subsection{Motion Forecasting Results} \label{sec:motion_forecasting}
\begin{figure*}[h]
\centering
    \includegraphics[width=0.8\textwidth]{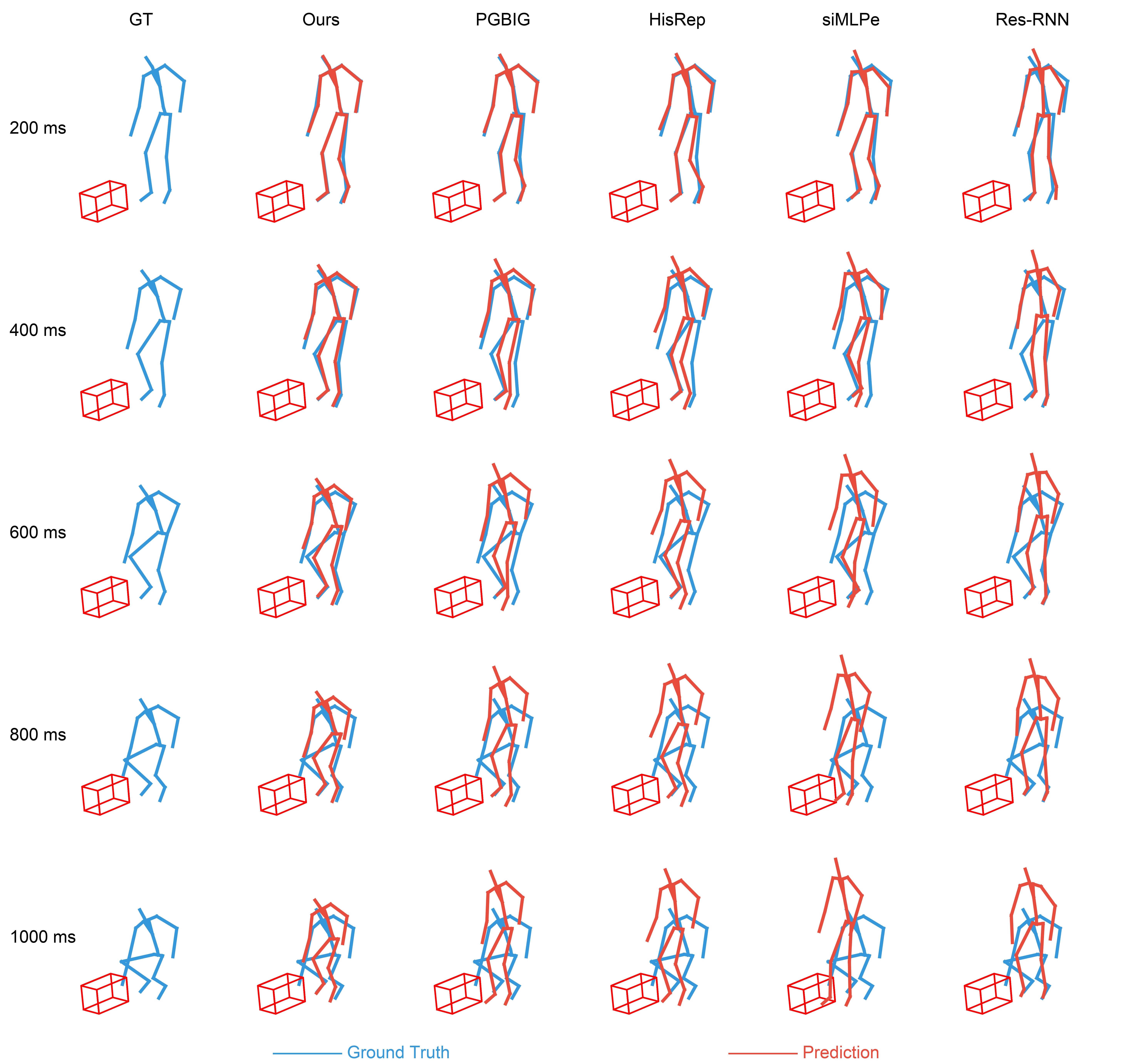}
    \caption{Visualisation of the prediction results from different methods on a sample of the ADT dataset~\cite{pan2023aria}. Our method can accurately predict the future body motion of \textit{squat down to touch an object} while prior methods that only use historical body poses fail to predict this motion.}
    \label{fig:result_adt}
\end{figure*}

\paragraph{Results on ADT}
\autoref{tab:global} summarises the motion forecasting performances of different methods on individual actions (\textit{ADT-work}, \textit{ADT-decoration}, \textit{ADT-meal}), and on all actions (\textit{ADT-all}).
The table shows the average MPJPE error (in millimeters) over all $30$ frames as well as the prediction errors for different future time horizons: $100$ ms, $200$ ms, \ldots, $1000$ ms.
As can be seen from the table, our method consistently outperforms the state-of-the-art methods on different actions (\textit{work}, \textit{decoration}, \textit{meal}, or \textit{all}).
For \textit{work}, \textit{decoration}, and \textit{meal} actions, our method achieves an average improvement of $8.5\%$ ($30.0$ \textit{vs.} $32.8$), $8.9\%$ ($46.9$ \textit{vs.} $51.5$), and $8.0\%$ ($38.9$ \textit{vs.} $42.3$) 
respectively in terms of MPJPE error.
For all actions, our method achieves an average improvement of $8.7\%$ ($37.7$ \textit{vs.} $41.3$) over the state of the art.
We also compared the prediction errors of different methods at future $100\text{-}1000$ ms respectively and observed that our method consistently outperforms prior methods at all the future time intervals.
We further performed a paired Wilcoxon signed-rank test to compare the performances of our method with the state of the art and the results validated that the differences between our method and the state of the art are statistically significant ($p<0.01$).
\autoref{fig:result_adt} shows an example of the predicted body poses from different methods on a sample of the ADT dataset.
On this sample, the user is going to squat down to touch an object on the ground.
We can see that our method can accurately predict this future body motion while prior methods that only use historical body poses fail to predict this motion.
See supplementary video for more prediction results. 

\begin{figure*}[h]
\centering
    \includegraphics[width=0.8\textwidth]{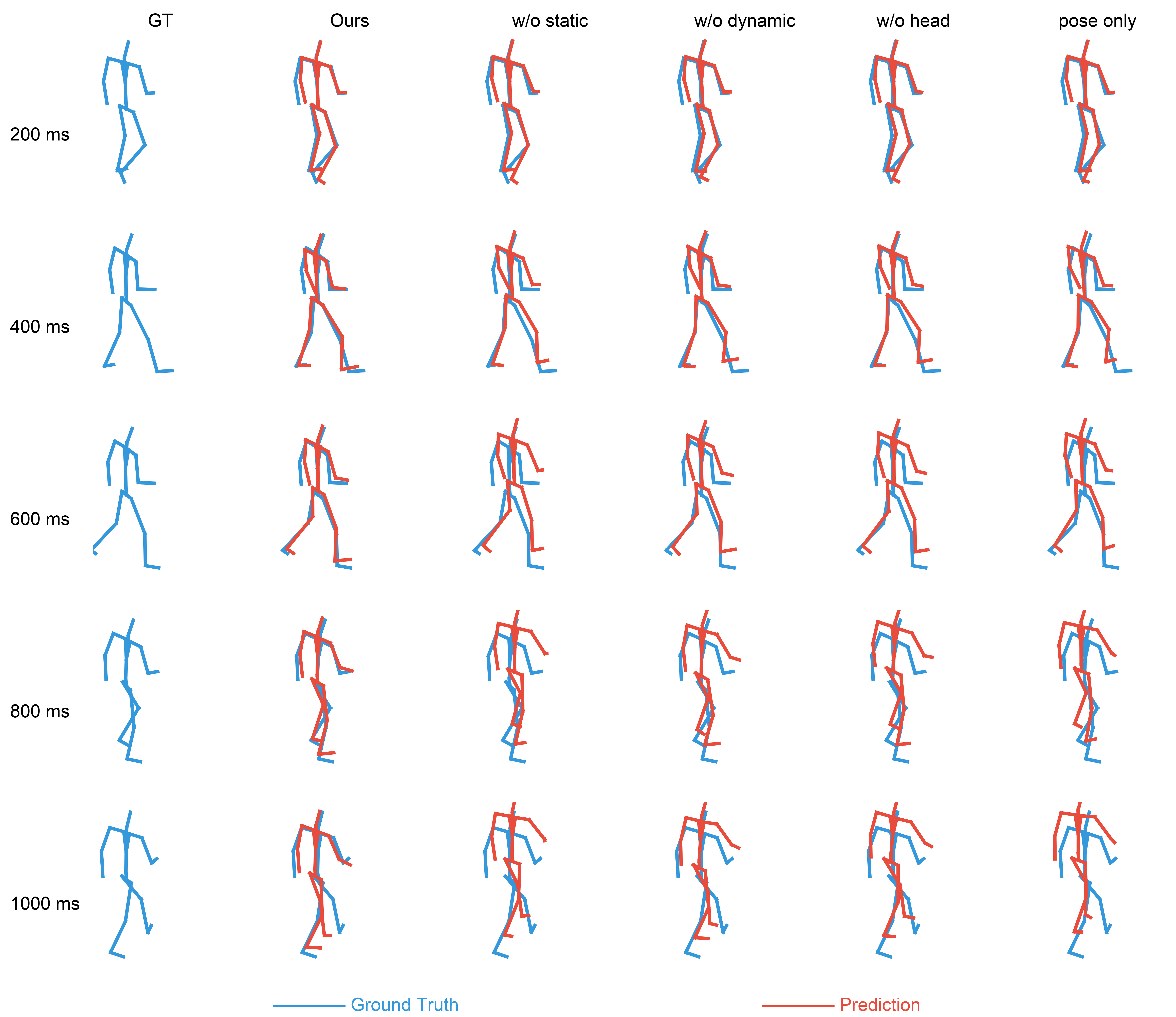}
    \caption{Visualisation of different ablated versions of our method on a sample of the MoGaze dataset~\cite{kratzer2020mogaze}. Our method consistently outperforms the ablated versions at different future time horizons. 
    }
    \label{fig:result_ablation}
\end{figure*}

\begin{table*}
        \caption{Mean per joint position errors (unit: millimeters) of different ablated versions of our method for motion forecasting on the MoGaze dataset. Best results are in bold. Our method significantly outperforms the ablated versions, validating the effectiveness of each component used in our method.}\label{tab:ablation_input}
	\centering
	\resizebox{0.9\textwidth}{!}{
	\begin{tabular}{cccccccccccc}
		\toprule
		Method & 100 ms & 200 ms & 300 ms & 400 ms & 500 ms & 600 ms & 700 ms & 800 ms & 900 ms & 1000 ms &Average\\ \hline
         w/o \textit{static} &13.8 &26.3 &39.7 &54.3 &70.2 &87.2 &105.3 &124.1 &143.4 &162.6 &77.3\\
         w/o \textit{dynamic} &13.8 &26.2 &39.6 &54.1 &69.9 &86.9 &105.0 &123.9 &143.2 &162.4	&77.1\\
         w/o \textit{static+dynamic} &13.9 &26.6 &40.0 &54.5 &70.5 &87.8 &106.0 &124.9 &144.3 &163.9 &77.8\\
         w/o \textit{head} &13.7 &26.2 &39.5 &54.2 &70.1 &87.2 &105.2 &124.1 &143.6 &163.0 &77.2\\
         w/o \textit{static+dynamic+head} &14.3 &26.9 &40.4 &55.0 &71.2 &88.8 &107.5	&126.9 &147.0 &167.3 &79.0\\
         \cline{1-12}
         Ours &\textbf{13.2} &\textbf{25.6} &\textbf{38.6} &\textbf{52.9} &\textbf{68.7} &\textbf{85.7} &\textbf{103.9} &\textbf{122.7}	&\textbf{142.0} &\textbf{161.3} &\textbf{76.1}\\
        \bottomrule                    
	\end{tabular}} 
 \end{table*}
 
\paragraph{Results on MoGaze}
The motion forecasting performances of different methods on individual actions (\textit{MoGaze-pick}, \textit{MoGaze-place}) and on all actions (\textit{MoGaze-all}) are summarised in \autoref{tab:global}.
The table shows the average MPJPE error (in millimeters) over all $30$ frames as well as the prediction errors for different future time horizons: $100$ ms, $200$ ms, \ldots, $1000$ ms.
We can see from the table that our method consistently outperforms the state-of-the-art methods on different actions (\textit{pick}, \textit{place}, and \textit{all}).
For \textit{pick} and \textit{place} actions, our method achieves an average improvement of $6.8\%$ ($72.7$ \textit{vs.} $78.0$) and $7.5\%$ ($82.6$ \textit{vs.} $89.3$) respectively in terms of MPJPE error.
For all actions, our method achieves an average improvement of $7.2\%$ ($76.1$ \textit{vs.} $82.0$) over the state of the art.
We also compared our method with prior methods at different future time horizons and validated that our method can achieve superior performances at all the future time intervals.
A paired Wilcoxon signed-rank test was further used to compare the performances of our method with the state of the art and the results validated that the differences between our method and the state of the art are statistically significant ($p<0.01$).

\paragraph{Using Only Historical Body Poses}
Considering that prior methods only used historical body poses as input, for a more fair comparison, we retrained our method using only body poses and tested on the ADT and MoGaze datasets.
\autoref{tab:global} summarises the motion forecasting performances on individual actions as well as on all the actions.
We can see that the ablated version of our method still outperforms prior methods on both the ADT and MoGaze datasets in terms of average performance as well as performances at different future time horizons.
For \textit{work}, \textit{decoration}, \textit{meal}, and \textit{all} actions on the ADT dataset, our method using only body poses achieves an average improvement of $5.8\%$ ($30.9$ \textit{vs.} $32.8$), $4.9\%$ ($49.0$ \textit{vs.} $51.5$), $5.4\%$ ($40.0$ \textit{vs.} $42.3$), and $5.3\%$ ($39.1$ \textit{vs.} $41.3$).
A paired Wilcoxon signed-rank test was further used to compare the performances of our method using only body poses with the state of the art and the results validated that the differences are statistically significant ($p<0.01$).
For \textit{pick}, \textit{place}, and \textit{all} actions on the MoGaze dataset, our method using only body poses achieves an average improvement of $4.1\%$ ($74.8$ \textit{vs.} $78.0$), $3.0\%$ ($86.6$ \textit{vs.} $89.3$), and $3.7\%$ ($79.0$ \textit{vs.} $82.0$).
The differences between our method using only body poses and the state of the art are statistically significant (paired Wilcoxon signed-rank test, $p<0.01$).
The above results validate the overall superiority of our model architecture.

\begin{table*}
        \caption{Mean per joint position errors (unit: millimeters) of our method using different numbers of scene objects, spatial nodes, pose residual GCN, and fuse residual GCN for motion forecasting on the MoGaze dataset. Best results are in bold.}\label{tab:ablation_architecture}
	\centering
	\resizebox{0.9\textwidth}{!}{
	\begin{tabular}{cccccccccccc}
		\toprule
		Method & 100 ms & 200 ms & 300 ms & 400 ms & 500 ms & 600 ms & 700 ms & 800 ms & 900 ms & 1000 ms &Average\\ \hline
        \textit{object} 0 &13.9 &26.6 &40.0 &54.5 &70.5 &87.8 &106.0 &124.9 &144.3 &163.9 &77.8\\
         \textit{object} 1 &13.5	&26.0 &39.3 &54.0 &69.9 &87.1 &105.1 &124.1 &143.5 &162.8 &77.1\\
         \textit{object} 2 (ours) 
         &\textbf{13.2} &\textbf{25.6} &\textbf{38.6} &\textbf{52.9} &\textbf{68.7} &\textbf{85.7} &\textbf{103.9} &\textbf{122.7}	&\textbf{142.0} &\textbf{161.3} &\textbf{76.1}\\
         \textit{object} 3 &13.6 &26.1 &39.3 &53.8 &69.6 &86.6 &104.7 &123.5 &142.8 &161.9 &76.8\\
         \textit{object} 4 &13.6 &26.2 &39.5 &54.1 &70.0 &87.1 &105.1 &123.8 &143.2 &162.5 &77.1\\
         \textit{object} 5 &13.8 &26.3 &39.7 &54.3 &70.3 &87.7 &106.0 &124.8 &144.4 &163.8 &77.7\\
         \midrule
        \textit{spatial node} 1 (no repeat) &14.0 &26.7 &40.1 &54.6 &70.3 &87.3 &105.3 &123.7 &142.7	&162.1 &77.3\\
        \textit{spatial node} 5 (ours) 
        &13.2 &25.6 &\textbf{38.6} &\textbf{52.9} &\textbf{68.7} &\textbf{85.7} &\textbf{103.9} &\textbf{122.7}	&\textbf{142.0} &\textbf{161.3} &\textbf{76.1}\\
        \textit{spatial node} 10 &13.3 &25.9 &39.2 &53.8 &69.8 &87.0 &105.2 &124.1 &143.5	&162.9 &77.0\\
        \textit{spatial node} 20 &13.1 &25.5 &38.7 &53.3 &69.3 &86.5 &104.8 &123.8 &143.3	&162.8 &76.7\\
        \textit{spatial node} 30 &\textbf{13.0} &\textbf{25.5} &38.8 &53.4 &69.3 &86.8 &105.1 &124.2 &143.6 &162.8 &76.8\\
        \midrule
        \textit{pose residual GCN} 0 &13.4 &25.8 &39.1 &53.6 &69.4 &86.5 &104.6 &123.3 &142.7 &161.9 &76.6\\
        \textit{pose residual GCN} 8 (ours) &\textbf{13.2} &\textbf{25.6} &\textbf{38.6} &\textbf{52.9} &\textbf{68.7} &\textbf{85.7} &\textbf{103.9} &\textbf{122.7}	&\textbf{142.0} &\textbf{161.3} &\textbf{76.1}\\
        \textit{pose residual GCN} 16 &13.4 &25.8 &39.2 &53.9	&70.0 &87.5 &105.8 &125.0 &144.6 &164.1 &77.5\\
        \midrule
        \textit{fuse residual GCN} 0 &16.1 &31.0 &46.0 &61.5 &78.0 &95.7 &114.7 &134.5 &155.3 &176.6 &85.1\\
        \textit{fuse residual GCN} 8 &14.3 &27.0 &40.3 &54.9 &70.6 &87.7 &105.8 &124.7 &144.1 &163.4 &77.8\\
        \textit{fuse residual GCN} 16 (ours) &\textbf{13.2} &\textbf{25.6} &\textbf{38.6} &\textbf{52.9} &\textbf{68.7} &\textbf{85.7} &\textbf{103.9} &\textbf{122.7}	&\textbf{142.0} &\textbf{161.3} &\textbf{76.1}\\
        \textit{fuse residual GCN} 32 &13.3 &25.8 &39.2 &53.8 &69.8 &87.1 &105.6 &124.8 &144.7 &164.4 &77.4\\
        \bottomrule                    
	\end{tabular}} 
\end{table*}

\paragraph{Time Costs and Model Size}
\autoref{tab:model_size} shows the time costs and model size of different methods.
\hl{We can see that our method is of medium size and is more efficient than prior methods in terms of test time per batch ($10$ ms), validating the usefulness of our method in real-time applications.}
The time costs were calculated on an NVIDIA Tesla V100 SXM2 32GB GPU with an Intel(R) Xeon(R) Platinum 8260 CPU @ 2.40GHz.

\begin{table}
        \caption{Time costs and model size of different methods.}\label{tab:model_size}
	\centering
	\resizebox{0.4\textwidth}{!}{
	\begin{tabular}{cccc}
		\toprule
		Method & Training (Per Batch) & Test (Per Batch) & Model Size\\ \hline
        \textit{Res-RNN}~\cite{martinez2017human} &37 ms &32 ms &3.41 M\\
        \textit{siMLPe}~\cite{guo2023back} &62 ms &36 ms &0.09 M\\
        \textit{HisRep}~\cite{mao2021multi} &56 ms &37 ms &3.38 M\\
        \textit{PGBIG}~\cite{ma2022progressively} &97 ms &39 ms &1.93 M\\
        Ours &59 ms	& 10 ms & 2.23 M\\
        \bottomrule                    
	\end{tabular}} 
\end{table}

\subsection{Ablation Study}

\paragraph{Scene Objects and Head Orientation}
In addition to historical body poses, our method also used features from head orientation and egocentric scene objects.
To evaluate the effectiveness of these inputs, we respectively removed \textit{static objects}, \textit{dynamic objects}, \textit{static and dynamic}, \textit{head orientation}, and \textit{static, dynamic, and head}, and retrained the ablated methods.
\autoref{tab:ablation_input} shows the motion forecasting results of these ablated methods on the MoGaze dataset.
We can see that our method consistently outperforms the ablated methods in terms of both average error and errors at different time horizons and the results are statistically significant (paired Wilcoxon signed-rank test, $p<0.01$).
The above results indicate that both scene objects and head orientation help improve our method's motion forecasting performance.
\autoref{fig:result_ablation} shows an example of the predicted body poses from different ablated versions of our method on the MoGaze dataset.
We can see that our method consistently outperforms the ablated versions at different future time horizons, validating the effectiveness of each component used in our method.

\paragraph{Scene Object Number}
In our method, we used the two dynamic objects and two static objects that are closest to the viewport centre as our input.
We further tested different number of scene objects and indicated the motion forecasting performances on MoGaze in \autoref{tab:ablation_architecture}.
The scene objects were selected according to their angular distances to the viewport centre (\autoref{sec:object_feature}).
For simplicity, the number of static and dynamic objects were kept the same.
We can see from \autoref{tab:ablation_architecture} that using two dynamic and static objects achieves the best performances and the differences between different number of scene objects are statistically significant (paired Wilcoxon signed-rank test, $p<0.01$).
We also noticed that using more objects does not boost the motion forecasting performance, probably because users are more likely to interact with the objects that are closer to the viewport centre than the objects in the peripheral region~\cite{sitzmann2018saliency, hu2019sgaze, hu2020dgaze} and thus adding information on peripheral objects cannot improve the performance.


\paragraph{Pose-object Graph}
In our pose-object graph, we respectively repeated the head and object features in the spatial domain to obtain five spatial nodes to enhance these features before fusing them with the pose features.
To evaluate the effectiveness of this strategy, we further tested different number of spatial nodes and indicated the motion forecasting performances on MoGaze in \autoref{tab:ablation_architecture}.
For simplicity, we used the same node number for head orientation, static objects, and dynamic objects.
We can see from \autoref{tab:ablation_architecture} that repeating the features to obtain five spatial nodes achieves the best performances and the differences between different number of spatial nodes are statistically significant (paired Wilcoxon signed-rank test, $p<0.01$).
We also noticed that repeating the features in the spatial domain (\textit{spatial node} $>1$ in \autoref{tab:ablation_architecture}) always performs better than no repeat (\textit{spatial node} $=1$), validating the effectiveness of the repeat strategy used in our pose-object fusion process.

\paragraph{Pose and Fuse Residual GCN}
In our method, we used pose residual GCNs to enhance the pose features and fuse residual GCNs to enhance the pose-object features.
To evaluate the effectiveness of these two GCN blocks, we respectively removed these two blocks or changed the number of GCNs and indicated the motion forecasting performances on MoGaze in \autoref{tab:ablation_architecture}.
We can see that using these two GCN blocks achieves significantly better performances than not using them (paired Wilcoxon signed-rank test, $p<0.01$), validating the effectiveness of these two blocks.
We also validated that using eight pose residual GCNs and $16$ fuse residual GCNs achieves the best motion forecasting performance.



\subsection{User Study} \label{sec:user_study}

The results in Section \ref{sec:motion_forecasting} have quantitatively validated the effectiveness of our method.
To further evaluate whether our method's improvements are significant in terms of human perception, we conducted a user study to compare our method with prior methods.

\subsubsection{Stimuli} 
We randomly selected $20$ motion forecasting samples from the ADT and MoGaze datasets ($10$ samples from each dataset) and used them as our stimuli.
Each sample consisted of $30$ frames of predictions (corresponding to future $1$ second) and was visualised as a short video.

\subsubsection{Participants}
We recruited 20 participants (10 males and 10 females, aged between
18 and 50 years, Mean=27.9, SD=6.8) to take part in our user study through university mailing lists and social networks.
All of the participants reported normal or corrected-to-normal vision.
The user study was approved by our university's ethical review board.

\subsubsection{Procedure}
We conducted our user study using a Google form.
During the study, the ground truth future motions and the predictions of different methods were displayed to the participants in parallel using a layout that is similar to \autoref{fig:result_adt}.
For simplicity, we only compared our method with PGBIG~\cite{ma2022progressively} and HisRep~\cite{mao2021multi} since they are the strongest baselines from the results in \autoref{tab:global}.
The names of different methods were hidden and the order of these methods were randomised.
The visualisation videos of the ground truth and different methods were set to loop automatically, allowing participants to observe them with no time limit.
\hl{Before the user study, participants were given detailed instructions (see the supplementary material) to get familiar with our experimental setting.}
During their observation, participants were required to rank different methods according to two criteria: \textit{precision} and \textit{realism}.
\begin{itemize}
\item \textit{Precision}: check different methods to see whether they \textit{align with the ground truth} and rank them based on your observation.
\item \textit{Realism}: check different methods to see whether they are \textit{physically plausible} and rank them based on your observation.
\end{itemize}
We collected the responses from all the participants for further analysis.

\subsubsection{Statistical Analysis}
The medians, means and standard deviations (SDs) of different methods' rankings are shown in Table \ref{tab:user_study}.
We can see that our method outperforms the state of the art in terms of both precision (Median: $1.0$ \textit{vs.} $2.0$, Mean: $1.2$ \textit{vs.} $2.3$) and realism (Median: $1.0$ \textit{vs.} $2.0$, Mean: $1.3$ \textit{vs.} $2.2$).
The results from a paired Wilcoxon signed-rank test validated that the differences between our method and the state of the art are statistically significant ($p<0.01$).
The above results demonstrate that our method achieves significantly better performances over prior methods in terms of human perception.

\begin{table}[t]
	\centering
        \caption{Statistical results of different methods' rankings in our user study. Best results are in bold. Rank 1 means best performance.}\label{tab:user_study}
	\resizebox{0.4\textwidth}{!}{
	\begin{tabular}{ccccc}
		\toprule
		 & & Ours & \textit{PGBIG}~\cite{ma2022progressively} & \textit{HisRep}~\cite{mao2021multi}\\ \hline
	\multirow{3}{*}{\textit{Precision}}
        & Median &\textbf{1.0} &2.0 &3.0 \\
        & Mean & \textbf{1.2} & 2.3 & 2.5\\
        & SD   & 0.5 & 0.6 & 0.6\\ 
        \midrule
	\multirow{3}{*}{\textit{Realism}}
        & Median &\textbf{1.0} &2.0 &2.0\\
        & Mean & \textbf{1.3} & 2.2 & 2.3\\
        & SD   & 0.6 & 0.7 & 0.7\\
        \bottomrule
	\end{tabular}}
\end{table}
\section{Discussion}

\paragraph{Significance of Our Method}
Our method consistently outperforms prior methods in terms of average
performance as well as performances at different time intervals (\autoref{tab:global} and \autoref{fig:result_adt}), and the differences between our method and the state of the art are statistically significant (\autoref{sec:motion_forecasting}, paired Wilcoxon signed-rank test, $p<0.01$).
\hl{The results from a user study further confirm that our improvements are significant in terms of human perception (\autoref{sec:user_study}), implying that our method can be more effective in real applications.}

\paragraph{Scene Objects for Motion Forecasting}
Our method combines past body poses with egocentric 3D object bounding boxes to forecast body motion in the future.
Extensive experiments validate that the 3D bounding box information of scene objects can significantly improve the performances of motion forecasting (\autoref{tab:ablation_input} and \autoref{fig:result_ablation}).
We also found that increasing the number of scene objects does not necessarily improve the motion forecasting performance (\autoref{tab:ablation_architecture}), revealing that users' body motions are mainly influenced by the objects that are closer to the viewport centre.
These results provide meaningful insights for developing future scene object-aware human motion forecasting methods.

\paragraph{Usability of Our Method}
\hl{Although our method requires additional information on the egocentric 3D object bounding boxes
, such information is readily available in VR/AR systems~\cite{park2008multiple, crivellaro2017robust, hu2021fixationnet, holoyolo} and can also be easily accessed in real-world environments by applying existing 3D object bounding box estimation methods~\cite{mousavian20173d}.}
In addition, even using only historical body poses as input, our method still achieves significantly better performances over prior methods (\autoref{tab:global}), further validating the usability of our method in real applications.

\hl{\paragraph{Head Orientation vs. Eye Gaze} We have tried to use eye gaze in our preliminary experiments and found that eye gaze performs worse than head orientation in terms of MPJPE error on ADT (38.6 vs. 37.7). This is probably because eye gaze is much more noisy than head orientation and thus degrades overall performance. Therefore, we opt to use head orientation in our architecture.}

\paragraph{Limitations}
Despite all the advances that we have made, we identified several limitations that we plan to address in future work.
\hl{First, to the best of our knowledge, the MoGaze and ADT datasets are the only public datasets that provide both full-body motion and information on 3D scene objects, thus unfortunately limiting the generalisability of our evaluation. In future work, we plan to assess our method for a broader range of activities and environments.}
In addition, our method is specifically designed for human-object interactions and may not work well for other situations such as human-human interactions.
How to adapt our method to other situations remains to be explored.
Finally, our method takes observed past body poses and scene objects as input while in real applications the input data may be incomplete due to tracking errors and may degrade the performance of our method.
How to deal with incomplete observations is worthy of further study.

\paragraph{Future Work}
Besides overcoming the above limitations, many potential avenues of future work exist.
First, it would be interesting to explore the effectiveness of other scene object-related information such as shape and colour for human motion forecasting.
In addition, we are also looking forward to adding some physical constraints for the predicted human poses to make them more physically plausible.
\hl{Furthermore, integrating our method into motion-related VR/AR applications is an interesting avenue of future work.}
Finally, adding prior knowledge on human intention during human-object interactions, e.g. the target object during a \textit{pick} activity, to our pipeline may further boost the motion forecasting performance.
\section{Conclusion}

In this work we proposed a novel method for human motion forecasting during human-object interactions that first uses an encoder-residual GCN and multi-layer perceptrons to extract features from past body poses and egocentric 3D object bounding boxes respectively, fuses these features into a pose-object graph, and applies a residual-decoder GCN to forecast future motion.
Through extensive experiments on two public datasets for motion forecasting at different time intervals we demonstrated that our method consistently outperformed several state-of-the-art methods by a large margin.
We also validated that our predictions were more precise and more realistic than prior methods through a user study.
As such, our work reveals the significant information content available in egocentric 3D object bounding boxes for human motion forecasting and informs future research on this promising research direction.

\acknowledgments{
This work was funded by the Deutsche Forschungsgemeinschaft (DFG, German Research Foundation) under Germany's Excellence Strategy -- EXC 2075 -- 390740016.
A. Bulling was funded by the European Research Council (ERC) under grant agreement 801708.
}

\bibliographystyle{abbrv-doi-hyperref}

\bibliography{references}

\end{document}